# A Novel Structured Natural Gradient Descent for Deep Learning

Weihua Liu[1], Xiabi Liu[1]

[1] Beijing Lab of Intelligent Information Technology, School of Computer, Beijing Institute of Technology, Beijing 100081, China

```
WHL: liuweihua@bit.edu.cn
XBL: liuxiabi@bit.edu.cn
```

Natural gradient descent (NGD) provided deep insights and powerful tools to deep neural networks. However the computation of Fisher information matrix becomes more and more difficult as the network structure turns large and complex. This paper proposes a new optimization method whose main idea is to accurately replace the natural gradient optimization by reconstructing the network. More specifically, we reconstruct the structure of the deep neural network, and optimize the new network using traditional gradient descent (GD). The reconstructed network achieves the effect of the optimization way with natural gradient descent. Experimental results show that our optimization method can accelerate the convergence of deep network models and achieve better performance than GD while sharing its computational simplicity.

## 1 Introduction

The traditional gradient descent (GD) for training deep neural network is popular because of its simplicity and high efficiency, but it only uses the first-order information and often encounter the pathological curvature problem [1] (the canyon area in the surface of the objective function, it makes the optimization only concentrate on a small set of parameters), which is easy to oscillate along the ridge of the canyon, the speed movement to the smallest direction is very slow, which makes the slow training process. As a second-order optimization method, natural gradient descent (NGD)[2] can not only converge quickly, but also effectively find the global minimum value, but when natural gradient descent is used to optimize the deep neural network, the computation cost of the Fisher information matrix and its inverse is too large. In order to perform calculations faster, we need to find natural gradient algorithms with low computational complexity and low storage requirements.

This paper proposes a structured natural gradient optimization method (SNGD) for learning deep neural networks. SNGD first reconfigures the parameter layer of the deep network by adding a new processing layer (named local Fisher layer); and then optimizes the reconstructed network model based on traditional GD, which is equivalent to the optimization of the original network using NGD, thus effectively reducing the computational complexity of NGD. With the introduction of the local Fisher layer, the curvature information of the loss function space can be captured, and an adjustment related to the spatial curvature is added to the original gradient direction, which ensures that there is a reasonable parameter change in each update during optimization, and improves the convergence speed of the parameters. We test the proposed approach on…

The main contributions of this paper are as follows:

1) By adding a new local Fisher layer to reconstruct the network, the relevant calculation of the global Fisher matrix is decomposed and finally transformed into the use of traditional GD for optimization to achieve the effect of NGD.

2) A new layer - local Fisher layer and its efficient implementation scheme are proposed. Through the introduction of the second-order information, the local Fisher layer considers the different attributes of different positions of the parameters, and adds constraints to the transformation of the model parameters, so that the gradient update can be carried out stably and quickly.

3) The proposed SNGD is applied to the classification network, and the experimental results show that, compared with the optimization of the traditional GD, the proposed SNGD optimization method has faster convergence speed, and the trained model also has better performance. In addition, extended experiments on other applications such as detection further verify the universal applicability of SNGD.

## 2 Related works

The work reported in this paper origins from natural gradient descent for training neural networks. The new proposed layer has a similar structure as the normalization layer. Thus, in this Section, we briefly review the two related topics: 1) natural gradient descent, 2) normalization methods.

**2.1 Natural gradient descent**

NGD was proposed by Amari et al[2] in 1998. The natural gradient is the fastest decreasing direction of the error defined in the parameter space in the Riemannian space. Amari's experiments show that the convergence speed of the algorithm is faster than the stochastic gradient descent (SGD). However, it is very difficult to implement NGD in deep neural networks, because the corresponding Fisher matrix is huge. In order to simplify the calculation of natural gradients, Bastian et al.[4] considered every two layers of the neural network as a model, and the resulting information matrix is called the block information matrix. Compared with natural gradient descent, the complexity of the natural gradient is greatly reduced, however the performance in the experiment is not ideal, and even unable to converge normally. The approximate curvature of Kronecker coefficients (K-FAC) proposed by Martens etc.[5] is also an effective method to implement the natural gradient algorithm, which is an approximate natural gradient algorithm for optimizing the cascading structure of neural networks. But the noise caused by approximate estimation will affect the accurate calculation of the Fisher matrix. Zhang et al[6] theoretically analyzed the convergence rate of NGD on nonlinear neural networks based on square error loss, determined two prerequisites for effective convergence after random initialization, and proved that although K-FAC is an approximate natural gradient method, it can converge to global optimization on the premise that it satisfies the prerequisite. George et al.[7] proposed a novel approximate natural gradient method, which is proved to be better and less expensive than K-FAC. The characteristic of George and other methods is that diagonal variance is used on the basis of Kronecker eigenfactors.

Bernacchia et al[8] derived the exact expression of the natural gradient in the deep linear network, which shows the morbid curvature similar to that in the nonlinear case. In view of the complexity of deep neural networks, Sun et al.[9] decomposed the neural network model into a series of local subsystems, on the basis of which the relative Fisher information metric was defined, which reduced the complexity of the optimization calculation of the whole network model. This local dynamic learning replaces the global learning of complex models, which makes it possible for complex deep neural networks to apply NGD, and one of its advantages is that the global Fisher information matrix is calculated accurately rather than approximately.

**2.2 Normalization method**

Sohl-Dickstein et al[10] believes that the effect of natural gradient on the parameter space is similar to the whitening of the signal. In this article, we use a similar normalization method to add additional layers to re-express the parameter space of each layer, so that the optimization can perform the gradient descent in the Riemann space.

As we all know, normalizing the input data prevents the network from constantly adapting to input changes and speeds up training. Normalization has been widely used in deep networks. For example, local response normalization is proposed in AlexNet[11]. BN is widely used in various networks, and it normalizes data along the batch dimension. But BN has no obvious effect when the batch dimension is small. In order to avoid the use of the batch dimension, some new normalization methods have been proposed. Layer Normalization (LN)[12] is normalized along the channel dimension. Instance normalization (IN)[13] perform BN-like operation, but only for each sample. Weight normalization (WN)[14] recommends standardizing the parameter weights of each layer rather than the features. Wu et al[15] proposed group normalization (GN) as a simple alternative to BN. GN divides the channels into groups and it is not affected by the batch size. Miyato etc.[16] proposed a new weight standardization technique, called spectrum normalization, is proposed, which can be stably used for training of GANs. Park etc.[17] proposed the spatial adaptive normalization to synthesize realistic images when the input semantic layout is given, which has advantages in terms of visual fidelity and alignment with the input layout.

# 3 Our method

In this section, the natural gradient descent (NGD) method is firstly introduced,

then the structured natural gradient descent method (SNGD) is derived. Finally, SNGD is extended to optimize deep neural networks, and the implementation skills are given.

**3.1 Natural gradient descent**

Natural gradient descent is a second-order optimization method for training statistical models. The update rules for the model parameter vector $w$ are as follows:

$$w^{(k+1)} = w^{(k)} - \alpha G^{-1} \frac{\partial \ell}{\partial w^{(k)}}, \quad (1)$$

where $\ell$ is the loss function, $\frac{\partial \ell}{\partial w^{(k)}}$ is the gradient of the loss function calculated in $k$-th iteration, $\alpha$ is the learning rate, $G$ is the Fisher matrix, which can be regarded as the Riemann metric on the statistical manifold and $G^{-1}$ is the inverse of the Fisher matrix.

Assuming that the probability distribution of the model is $p(x)$, the Fisher matrix $G$ is given by:

$$G = \mathbb{E}_{p(x)}[\nabla_w \ell (\nabla_w \ell)^T] = \text{Cov}_{p(x)}(g). \quad (2)$$

The Fisher information matrix (positive definite) defines the local curvature of the model distribution space. The natural gradient adds a curvature-related adjustment to the original gradient direction. Therefore, the natural gradient descent is the method in which the parameter vector $w$ defined in the Riemann space and in accordance with the fastest descent method in the model distribution space.

**3.2 Structured natural gradient descent**

In this section, we first discuss the relationship between the natural gradient descent (NGD) and … (少了一方) and give a lemma; furthermore, the proposed SNGD method is deduced.

**Lemma 1**: *If there is a relationship between the models' parameters: $w' = w \cdot G^{\frac{1}{2}}$, then optimize parameter $w'$ based on GD is equivalent to optimize parameter $w$ based on NGD.*

Here we give the derivation of the above lemma:

> The update rule of natural gradient descent (formula (1)) is simplified to:
>
> $$w = w - G^{-1} \cdot \frac{\partial l}{\partial w}$$
>
> Both sides of the above equation are multiplied by $G^{\frac{1}{2}}$ and get:
>
> $$w \cdot G^{\frac{1}{2}} = w \cdot G^{\frac{1}{2}} - G^{-1} \cdot G^{\frac{1}{2}} \cdot \frac{\partial l}{\partial w}$$
>
> $$= w \cdot G^{\frac{1}{2}} - G^{-\frac{1}{2}} \cdot \frac{\partial l}{\partial w}$$
>
> An equivalent transformation of $\frac{\partial l}{\partial w}$, and get:
>
> $$w \cdot G^{\frac{1}{2}} = w \cdot G^{\frac{1}{2}} - G^{-\frac{1}{2}} \cdot \frac{\partial l}{\partial (w \cdot G^{\frac{1}{2}})} \cdot \frac{\partial (w \cdot G^{\frac{1}{2}})}{\partial w} \quad (1)$$
>
> After taking the derivative $\frac{\partial (w \cdot G^{\frac{1}{2}})}{\partial w}$ of the above formula (1) and get:
>
> $$w \cdot G^{\frac{1}{2}} = w \cdot G^{\frac{1}{2}} - G^{-\frac{1}{2}} \cdot G^{\frac{1}{2}} \frac{\partial l}{\partial (w \cdot G^{\frac{1}{2}})}$$
>
> $$= w \cdot G^{\frac{1}{2}} - \frac{\partial l}{\partial (w \cdot G^{\frac{1}{2}})}$$
>
> If replaced $w' = w \cdot G^{\frac{1}{2}}$ in the above formula, the following update rule of the gradient descent can be obtained:
>
> $$w' = w' - \frac{\partial l}{\partial w'} \quad (2)$$
>
> **Conclusion**: If there is a relationship: $w' = w \cdot G^{\frac{1}{2}}$, then optimize parameter $w'$ using GD is equivalent to optimize parameter $w$ using NGD.

According to Lemma 1 in Section 3.2, we can transform the model parameter $w$ to:

$$w = w' \cdot G^{-\frac{1}{2}}, \quad (3)$$

where $w'$ is the new parameter form, and $G^{-1/2}$ can be represented by an additional calculation operation.

Then, using NGD to optimize $w$ is equivalent to using GD to optimize $w'$. The gradient descent update rule can be simplified as follows:

$$w' = w' - \frac{\partial l}{\partial w'}. \quad (4)$$

According to the above analysis, the proposed structured natural gradient descent (SNGD) is derived: First transform the original parameters $w$ using the equation $w = w' \cdot G^{-1/2}$ by a structural decomposition, then optimize $w'$ by the traditional GD, which will lead to the same effect as optimizing $w$ by NGD.

**3.3 Learning deep neural networks**

Training deep neural networks by NGD, the calculation efficiency of the global Fisher matrix is still too large. In this paper, we use the idea of decomposition to decompose the deep network into subsystems (layers) in a hierarchical way, and finally decompose the calculation of the global Fisher information matrix of the network to the calculation of only the local Fisher matrix (related to a single subsystem). This decomposition method is similar to Sun et al.[9] (in Seciton 2.1). The local Fisher information matrix in each layer is calculated as follows:

$$G = E(v_f)E(xx^\top), \quad (5)$$

where $x$ represents the input of one layer, $f$ is the activation function of this layer, and $v_f$ is the derivative of the activation function, its meaning is the effective learning area of the neuron.

The hierarchical idea is used to decompose the network. Suppose that the parameter vector $w$ of a network layer is transformed by formula (3), where $G^{-\frac{1}{2}}$ can be represented by an additional layer of the network, this additional layer can be regarded as a new normalized layer. In this way, after we restructure the network and adjust the structure of the network through the new normalization layer, although the traditional gradient descent is used to optimize it, it has the same advantages as the natural gradient descent.

Let's take a simple multi-layer perceptron (MLP) shown in Figure 2 as an example to illustrate our method, where $f$ is the activation function, such as linear rectification function (ReLU), sigmoid function (Sigmoid), and so on. The default assumption here is that the activation function is Sigmoid. For each neuron of the Sigmoid activation

function, there are

$$v_{\text{Sigmoid}} = \text{Sigmoid}(w^\top x)\,[1 - \text{Sigmoid}(w^\top x)]. \quad (6)$$

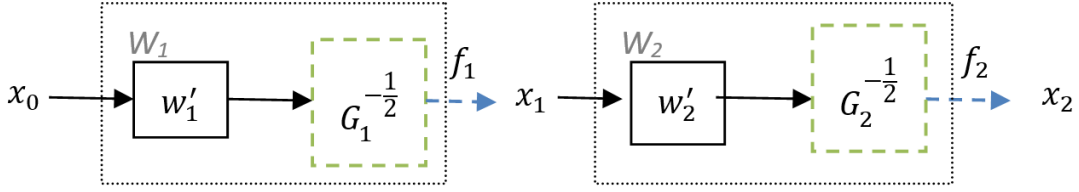

Figure 2 A simple MLP network with only two layers

Therefore, according to the above-mentioned hierarchical structured natural gradient optimization idea, the two-layer MLP network in Figure 1 can be optimized in the following way:

**Algorithm 1**. SNGD on a two-layer MLP network

---

Initialization $G_1^{-\frac{1}{2}}$ and $G_2^{-\frac{1}{2}}$ are the identity matrix,

Step 1: Calculation $G_1'^{-\frac{1}{2}}$

    1.1 First calculate $G_1'$, according to formula (5), $G_1' = E(V_{f_1}^2)E(x_0 x_0^T)$,

    where $V_{f_1} = \text{Sigmoid}(w_1'^T G_1^{-\frac{1}{2}} x_0)[1 - \text{Sigmoid}(w_1'^T G_1^{-\frac{1}{2}} x_0)]$

    1.2 Use the efficient method described in section 3.5 below to calculate $G_1'^{-\frac{1}{2}}$

Step 2: Update and replace $G_1^{-\frac{1}{2}}$ (after the replacement, $G_1^{-\frac{1}{2}}$ is a non-identity matrix)

$$G_1^{-\frac{1}{2}} = G_1'^{-\frac{1}{2}}$$

Step 3: Forward output $x_1$

$$x_1 = f_1\left(w_1' G_1^{-\frac{1}{2}} x_0\right).$$

Step 4: Calculation $G_2'^{-\frac{1}{2}}$

    4.1 Calculate $G_2'$ firstly, according to formula (5), $G_2' = E(V_{f_2}^2)E(x_1 x_1^T)$

    where $V_{f_2} = \text{Sigmoid}(w_2'^T G_2^{-\frac{1}{2}} x_1)[1 - \text{Sigmoid}(w_2'^T G_2^{-\frac{1}{2}} x_1)]$

    4.2 Calculate $G_2'^{-\frac{1}{2}}$

Step 5: Update and replace $G_2^{-\frac{1}{2}}$ as follows

$$G_2^{-\frac{1}{2}} = G_2'^{-\frac{1}{2}}$$

Step 6: Continue to forward output $x_2$

$$x_2 = f_2\left(w_2' G_2^{-\frac{1}{2}} x_1\right)$$

Step 7: Use traditional gradient descent to optimize $w_1'$ and $w_2'$

(Note that $G_1^{-\frac{1}{2}}$ and $G_2^{-\frac{1}{2}}$ does not participate in the backward propagation, as shown in the green box in Figure 2)

Iterate steps 1 to 7 in this way until convergence or termination of training.

For a complex multi-layer deep neural network, we first divide the network into some subsystem layers (composed of a linear transformation layer with a parameter vector plus a nonlinear activation function), and add a normalized layer to each layer to calculate the negative square root of the local Fisher matrix. The optimization method is the same as the optimization process of the two-layer MLP network described above.

The above analysis shows that SNGD reconstructs the deep neural network by adding a new "normalized layer" (local Fisher layer), and continues to use the traditional gradient descent for optimization. This is similar to the idea of BN, but the calculation of the normalization layer is different. From the formula (5), the calculation of the local Fisher layer not only includes regularizing the input data (through $E(xx^\top)$), but also considering the difference of transformation (through $E(v_f)$ select effective neurons). Through the regularization and nonlinear transformation of our "normalization layer", we can not only maintain the stability of the data distribution, but also skillfully provide the curvature signal of the loss space. In essence, it depends on the local Fisher layer to adjust the gradient direction of the parameters, and the natural gradient has been proved to accelerate the convergence. Therefore, our approach not only accelerates the convergence of the network, but also plays a role of regularization.

### 3.4 Implementation tricks

From the above calculation process, we can see that the calculation is mainly focused on solving the negative square root of the local Fisher information matrix. We

optimize the calculation through the following two strategies to further reduce the computational cost:

**1）Calculation $xx^T$**

From formula (5), it can be seen that the calculation of the local Fisher matrix $G$ is the product of $E(v_f)$ and $E(xx^T)$, and the main calculation cost is on the calculation of $E(xx^T)$. And $xx^T$ in the neural network corresponds to a Gram matrix, which is a description of the distribution characteristics of the vector itself. Taking the input is 2D image as an example, assuming that the batch size is $N$ and the size of feature map $x$ by a certain network layer is $(N, C, H, W)$. The method of calculating the Gram matrix corresponding to the feature map $x$ is as follows: First, the input is flattened and then transposed to obtain two tensors. one tensor' size is $(N, C, H * W)$, and the other's size is $(N, H * W, C)$; then the matrix multiplication is performed on the first and second dimensions of these two tensors. The essence is to do the inner product for each channel of each sample, that is, to find each vector $H * W$ 's Gram and finally a Gram matrix (the size is $(N, C, C)$) is obtained.

**2) Find the negative square root of the matrix**

Instead of accurately calculating the square root of the matrix, we use Newton's method to iteratively solve the square root $Z$ of $A$ in equation 7:

$$F(Z) = Z^2 - A = 0 \quad (7)$$

Denman-Beavers iteration method is used here[18], Given $Y_0 = A$ and $Z_0 = I$, where $I$ is the identity matrix, the iterative operation is defined as follows:

$$Y_{k+1} = \frac{1}{2}(Y_k + Z_k^{-1}), Z_{k+1} = \frac{1}{2}(Z_k + Y_k^{-1}), \quad (8)$$

The matrice $Y_k$ and $Z_k$ can quickly converge to $A^{1/2}$ and $A^{-1/2}$ (in the experiment, it was found that it takes about 20 iterations)

In order to further improve the calculation speed, we refer to the iterative method[19], the iterative formula (8) is further modified to avoid the inverse operation. The optimization iterative operation is as follows:

$$Y_{k+1} = \frac{1}{2}Y_k(3I - Z_k Y_k), Z_{k+1} = \frac{1}{2}(3I - Z_k Y_k)Z_k. \quad (9)$$

In this way, the negative square root of the matrix can be obtained only by matrix multiplication, and its calculation speed is much faster than that of using singular value decomposition (SVD), and has an order of magnitude improvement. At this time, the number of iterations is also reduced, and in the experiment, it is found that only a few iterations are needed.

## 4 Experiments

In order to verify the effectiveness of the method in this paper, we first select the Handwritten digit classification and lung nodule classification to verify our method. Furthermore, extended experiments on other applications such as detection further verify the universal applicability of SNGD.

### 4.1 Image Classification

#### 4.1.1 Handwritten digit classification

Classification of handwritten digits in MNIST[20]Above, we compared the SNGD method with the SGD method. The sizes of the MNIST training set, validation set, and test set are 50,000, 10,000, and 10,000 images, respectively, and each sample is a 28×28 grayscale image.

Here, a 4-layer multi-layer perceptron (MLP) network is used to classify MNIST numbers. The network architecture shape is 784-80-80-80-10, which uses the ReLU activation function, and finally uses a Softmax layer for classification output. The loss function is a cross-entropy loss function based on L2 regularization. In order to make a fair comparison, in different experiments, except that the optimization method is different, the network architecture, batch size (50) and L2 regular term (10-3) and other settings used are exactly the same.

Figure 3 shows the learning curve of SNGD and SGD when training on the MNIST application. It can be seen from Figure 3 that the convergence speed of SNGD is significantly better than that of SGD, especially before 40 epochs. At the same time, the classification accuracy of the model after SNGD optimization will be better. After about 60 epochs, the accuracy of the verification set will continue to be greater than that of the SGD method, and the final classification accuracy will increase from 97.62%

to 97.79%.

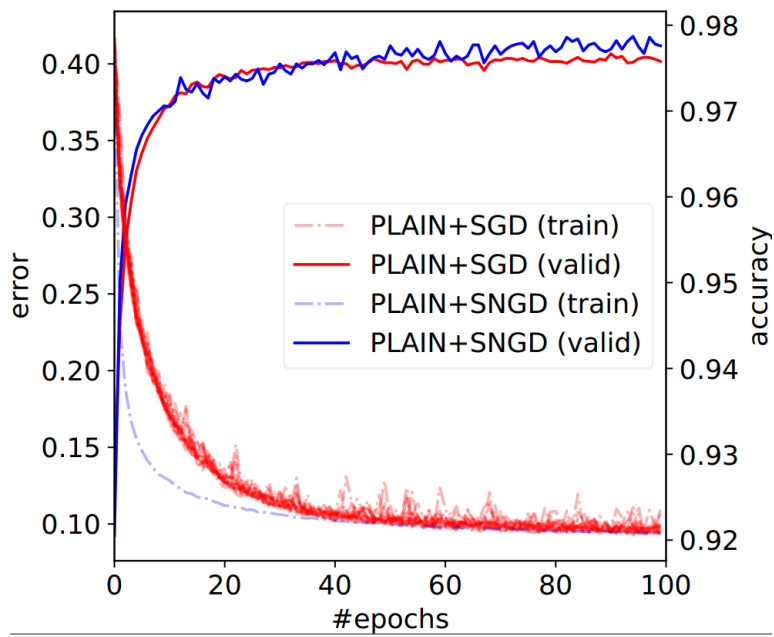

Figure 3 Comparison of SNGD and SGD optimization methods

### 4.1.2 Lung Nodule Classification

We perform the classification task to identify the correct nodule based on the 641,822 lung nodule candidate regions extracted from the 888 CT scans of LIDC-IDRI provided in the data LUNA16 challenge. We use SGD or SNGD based on the momentum set to 0.9 to train multi-level context 3D ConvNets (Dou et al., 2016) to achieve classification. The final performance is also measured by the Competitive Performance Index (CPM). First, we record the detection sensitivity, which is the number of identified true nodules divided by truly labeled nodules and the number of false positives per scan (FPs/scan). Then, CPM is defined as the average sensitivity of seven predefined FPs/scans: 1/8, 1/4, 1/2, 1, 2, 4, and 8. It should be noted that there is no BN layer in the network structure of ConvNets. Table 1 show that the performance of our SNGD is better than that of SGD. This result proves the value of our method relative to SGD once again.

Table 1 Comparison of FROC performance under nodule classification using different optimization methods in ConvNets

| Method | 1/8 | 1/4 | 1/2 | 1 | 2 | 4 | 8 | CPM |
|---|---|---|---|---|---|---|---|---|
| SGD | 0.677 | 0.834 | 0.927 | 0.967 | 0.979 | 0.980 | 0.981 | 0.906 |
| SNGD | 0.701 | 0.855 | 0.941 | 0.976 | 0.982 | 0.983 | 0.983 | 0.917 |

## 4.2 Object Detection

In order to determine whether the proposed optimization method has universal applicability, we further validated it on the lung nodule detection task. The experimental results also verify the effectiveness of our SNGD optimization method again, as described below.

**4.2.1 Lung nodule detection task**

This section evaluates and compares the performance of nodule detection when using SGD, SGD+BN, and SNGD to optimize PiaNet. Due to GPU memory limitations, the batch size of the BN layer can only be set to 2 during training. All experiments were trained and evaluated under the LUNA16 data set, and the FROC curve (sensitivity accuracy under different false positive rates) was used to evaluate and compare the performance of the model trained under different optimization methods. For detailed information about the LUNA16 data set and FROC curve, please refer to PiaNet.

Figure 3 shows the loss learning curve of PiaNet under the three optimization methods. It can be seen that the loss of the network under SNGD drops quite quickly. In the case of about 20 epochs, the loss value can be reduced to about 0.13, while the SGD or SGD+BN method requires at least 100 epochs. This indicates that the SNGD method proposed in this paper converges faster than SGD or SGD+BN.

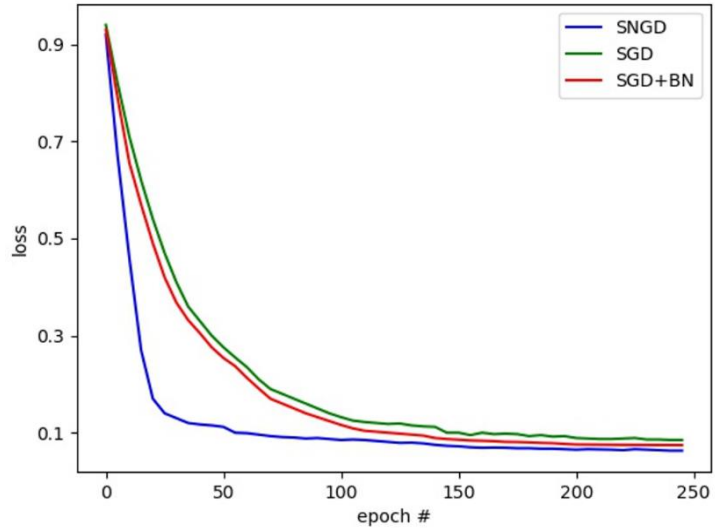

Figure 3 PiaNet loss curve under different optimization methods

Table 2 shows the corresponding FROC performance under different optimization methods. It can be seen that the performance of the SNGD optimization method is significantly better than the performance of the original PiaNet (BN size is 2) using SGD optimization. As shown in Table 2, the new optimization algorithm not only improves the sensitivity, but also improves the sensitivity under the same sensitivity.The false positive rate has also been reduced, and its overall performance index CPM value has been increased from 0.910 to 0.927. The above results prove that our method has obvious advantages in training complex detection models. In addition, we also found that the accuracy of the SGD+BN optimization method is only slightly improved compared to the SGD optimization method. This result shows that the effect of BN is quite limited when the batch size is relatively small.

Table 2 Comparison of FROC performance of PiaNet using different optimization methods on the LUNA16 data set

| Optimization | 1/8 | 1/4 | 1/2 | 1 | 2 | 4 | 8 | CPM |
| --- | --- | --- | --- | --- | --- | --- | --- | --- |
| **SGD,BN=2** | 0.738 | 0.875 | 0.923 | 0.955 | 0.96 | 0.961 | 0.961 | 0.910 |
| **SGD,withoutBN** | 0.725 | 0.874 | 0.918 | 0.949 | 0.957 | 0.958 | 0.959 | 0.906 |
| **SNGD** | **0.806** | **0.893** | **0.927** | **0.958** | **0.966** | **0.971** | **0.972** | **0.927** |

## 6 Conclusions

In this article, we introduce a novel optimization framework for deep neural networks. We propose an optimization method that realizes natural gradient descent by reconstructing the network layer in a deep neural network, referred to as SNGD, which can accelerate the network convergence and improve the optimization effect. Experimental results show that compared with the original traditional GD optimization methods, SNGD has achieved faster optimization speed and better model accuracy in handwritten digit recognition and lung nodule classification tasks. In addition, our SNGD method also shows better results than SGD and faster convergence speed in detection task, showing its universal applicability. we believe our method can provide a better optimization method for deep network learning.